\documentclass[10pt,twocolumn,letterpaper]{article}

\usepackage[final]{iccv}              
\usepackage[accsupp]{axessibility}  

\definecolor{iccvblue}{rgb}{0.21,0.49,0.74}
\usepackage[pagebackref,breaklinks,colorlinks,allcolors=iccvblue]{hyperref}
\def\paperID{8872} 
\def\confName{ICCV}
\def\confYear{2025}

\title{Everything is a Video: Unifying Modalities through Next-Frame Prediction}

\author{G. Thomas Hudson\qquad Dean Slack\qquad Thomas Winterbottom\qquad Jamie Sterling\\\qquad Chenghao Xiao\qquad Junjie Shentu\qquad Noura Al Moubayed\\\\Department of Computer Sciences, Durham University\\{\tt\small g.t.hudson@durham.ac.uk}}

\begin{document}

\twocolumn[{
\maketitle
\begin{center}
    \vspace{-2em}
    \captionsetup{type=figure}
        \includegraphics[width=0.19\linewidth]{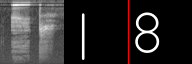}
        \includegraphics[width=0.19\linewidth]{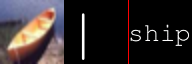}
        \includegraphics[width=0.82\linewidth]{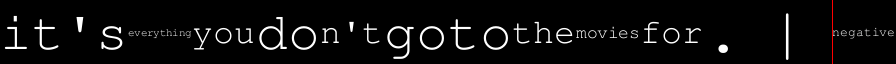}
        \includegraphics[width=0.82\linewidth]{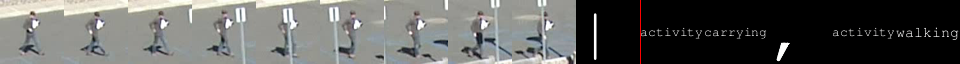}
        \includegraphics[width=0.82\linewidth]{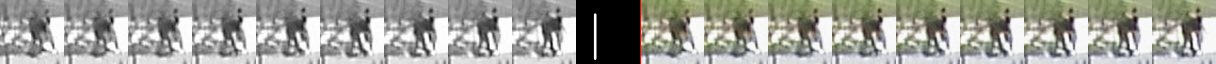}
        \includegraphics[width=0.82\linewidth]{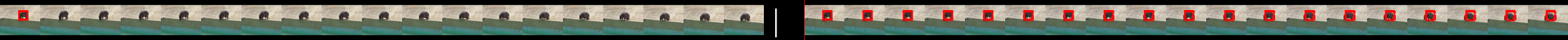}
       \includegraphics[width=0.83\linewidth]{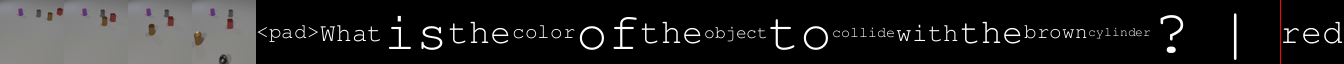}
    \vspace{-0.5em}
    \captionof{figure}{
    Sample outputs from our next frame prediction model across multiple modalities. 
    The model receives frames to the left of the red line as input and everything right of the red line has been generated by the model.
    \label{fig:sample-outputs}}
\end{center}
}]

\maketitle

\begin{minipage}{\columnwidth}\vspace{0.2em}
\begin{abstract}

Multimodal learning, which involves integrating information from various modalities such as text, images, audio, and video, is pivotal for numerous complex tasks like visual question answering, cross-modal retrieval, and caption generation. Traditional approaches rely on modality-specific encoders and late fusion techniques, which can hinder flexibility when adapting to new tasks or modalities. To address these limitations, we introduce a novel framework that extends the concept of task reformulation beyond natural language processing (NLP) to multimodal learning. We reformulate diverse multimodal tasks into a unified next-frame prediction problem, allowing a single model to handle different modalities without modality-specific components. This method treats all inputs and outputs as sequential frames in a video, enabling seamless integration of modalities and effective knowledge transfer across tasks. Our approach is evaluated on a range of tasks, including text-to-text, image-to-text, video-to-video, video-to-text, and audio-to-text, demonstrating the model's ability to generalize across modalities. We show that task reformulation can significantly simplify multimodal model design across various tasks, laying the groundwork for more generalized multimodal foundation models.

\end{abstract}
\end{minipage}\vspace{-1em}

\section{Introduction}
\label{sec:intro}

Many tasks we seek to solve are multimodal. The ability to process and integrate information from multiple modalities such as text, image, audio, and video, have proven crucial for a range of complex tasks, including cross-modal retrieval \cite{wang2024multimodal}, visual question answering \cite{wu2017visual, zou2020survey, lu2023multi}, and caption generation \cite{Agarwal2023FromMT}. 
The ability to understand and reason about information from different sensory inputs simultaneously mirrors human perception and cognition, making multimodal models an essential component in advancing artificial intelligence towards more generalized and versatile capabilities.

The predominant approach to building multimodal models involves using separate encoders for each modality. For example, in vision-and-language models, a convolutional neural network (CNN) or a vision transformer (ViT) \cite{dosovitskiy2021an} might be used to encode images, while a recurrent neural network (RNN) or transformer is used to encode text. These modality-specific encoders are then fused at a later stage, typically via concatenation or attention mechanisms, to enable the model to perform joint reasoning over the encoded representations. While effective, this paradigm has inherent limitations. It necessitates careful design choices for each modality and often struggles with scalability and flexibility, especially when extending to new combinations of modalities or tasks not seen during training. 

To address these challenges, we propose a novel approach centered around task reformulation. Task reformulation is a well-established paradigm in natural language processing (NLP) that facilitates multitask learning by transforming diverse tasks into a common format \cite{mccann2018natural}. A prominent example is prompt-based learning, where NLP tasks are reimagined as instructions or prompts to a large language model (LLM). This paradigm allows a wide range of NLP tasks, such as translation, summarization, and question answering, to be represented uniformly in a chat message format \cite{openai2024gpt4technicalreport}. By doing so, it leverages the generalization capabilities of large pre-trained models, enabling them to perform multiple tasks with minimal task-specific customization.

In this work, we extend the concept of task reformulation beyond NLP to the realm of multimodal learning. Specifically, we propose a framework in which tasks across various modalities are reformulated as a unified next-frame prediction problem. This approach simplifies the design of multimodal models by enabling them to handle diverse input types — text, images, audio, or video using a single, coherent mechanism. By representing all tasks as next-frame prediction, we create a shared interface for the model to learn and reason about information across modalities. This not only facilitates the integration of new modalities with minimal effort but also enhances the model’s ability to transfer knowledge across different tasks and domains. Ultimately, just as LLMs function as foundation models for NLP, we intend this framing to be a foundation model across all modalities.

The main contributions of this paper are as follows:

\begin{itemize}
    \item We propose a task reformulation method and show how to represent a range of multimodal tasks as next-frame prediction.
    \item We explore how this reformulation allows a single transformer-based model to solve text, image, video, and audio processing tasks without any modality-specific encoders.
    \item Experimental results show our model achieves comparable performance to single-task models which haven't been pretrained on additional data.
\end{itemize}

\section{Related work}
\label{sec:related}

\paragraph{Multimodal Learning}

With the rise of transformers, LLMs, and very large scale pretraining, new paradigms for harmonising different modalities have been pushed to the forefront of multimodal learning. UNITER \cite{Chen2019UNITERLU} uses a single universal transformer to process both image and text inputs into a join embedding space, with image representations prepended to the beginning of the input embeddings to the transformer. However, the image inputs are first processed by an RNN to obtain an embedding such that it is compatible with the textual inputs. Our framework in contrast has no need to preprocess input modalities differently as they are unified into a single visual input space theoretically capable of fully representing all multimodal interactions. \citet{Baevski2022data2vecAG} propose data2vec, a framework for instilling features from multiple modalities into a single latent representation. Where data2vec seeks to process raw text, images, and audio using the same method, our work here reformulates the task inputs from multiple modalities into the \textit{same visual input medium}.  UniT \cite{Hu2021UniTMM}, a multimodal multitask transformer uses separate encoders for each modalitiy followed by a shared \textit{decoder}. Our work uses a single visual encoder for our unified visual input paradigm. The most powerful state-of-the-art multimodal models such as FLAVA \cite{DBLP:journals/corr/abs-2112-04482} and GPT-4 vision \cite{openai2024gpt4technicalreport} integrate vision inputs via a patchwise process pipeline through a visual transformer fully integrated into the training pipeline. Both the architecture design and scale at which these foundation models are trained yield revolutionary performance at a variety of open-ended multimodal tasks. Nonetheless, though these foundation model are the closest that benchmarks have practically come to seamlessly integrating images with visual inputs, fundamentally the input modalities are handled differently by the model. Our work here bridges this remaining gap in processing of separate modalities. We reformulate typically spura-visual tasks into a visual input domain with \textit{no loss of information} \textit{i.e.} text and audio can be fully represented in images. This reformulation allows us to ---for the first time--- explore the generative pretraining capacity of multimodal information in one truly unified representation.

\paragraph{Task Reformulation}

Task reformulation involves allowing a single model to solve multiple tasks be converting them into a single `supertask'. This technique is common in NLP where a range of tasks such as sentiment classification, and coreference resolution have been reformulated as span-extraction \cite{keskar2019unifying}, and question answering \cite{mccann2018natural} among others.

The recent trend of prompt-based learning uses language modelling on prompt-answer sequences as the `supertask' \cite{openai2024gpt4technicalreport}. In this paradigm, the model is trained to follow an instruction describing the task and provide a response. In this way, any NLP task can be reformulated as the language modelling `supertask'. Similar ideas have been explored in computer vision, notably the Transframer \cite{nash2022transframer} which aimed to unify visual tasks. In our work, we extend this concept, exploring how next-frame prediction can be used as a `supertask' to unite multiple modalities.

\paragraph{Next-frame Prediction}

CNN-based deep learning architectures for the autoregressive generation of future video frames have been steadily improving over the last decade \cite{Ranzato2014VideoM, Amersfoort2017TransformationBasedMO, Yilmaz2021DFPNDF, Seo2023ImplicitSA}.  Transformer-based models trained at scale such as VideoGPT \cite{yan2021videogpt} have substantially improved the performance of quality and fidelity of long and short term future frame predictions. Modern diffusion-based frame prediction models are now able to generate photorealistic outputs \cite{Gupta2023PhotorealisticVG}. The aim of our work is \textit{not} a new model architecture. Our model is inspired from state-of-the-art architectures to serve as an appropriately powerful benchmark for our analysis on our benchmark datasets.

\paragraph{Visual Representation of Text}

Inspired by the success of unsupervised next-word prediction with language models, learning the next pixel of images was proposed \cite{chen2020generative} and shown able to learn image representations. Learning textual semantics with vision models, by rendering texts into images, has drawn more and more attention \cite{rust2022language,tai2024pixar,gao2024improving} due to the drawbacks of tokenization in traditional language models and limitations in cross-lingual transferability. \citet{xiao2024pixel} show that vision models pretrained on rendered images have stronger robustness on typo and word-order shuffling perturbations, and the representations of them display better isotropy on out-of-distribution (OOD) languages. In this work, we extend this approach when reformulating NLP tasks to video as well as other modalities which output text.

\section{Method}

Our proposed framework consists of two key components: (1) Methods of reformulating a diverse range of input and output modalities into the single task of next-frame prediction; (2) A pure transformer-based model architecture. We introduce these components in the following sections.

\subsection{Reformulation}

The datasets we use are carefully selected to cover a diverse range of input and output modalities, illustrating the versatility of our approach in reformulating various tasks as next-frame prediction problems. Our selection spans tasks from text and image classification to more complex audio, video, and multimodal tasks, enabling us to evaluate the model's ability to generalize across different data types and task requirements. By converting each task into a uniform format—a $64 \times 64$ RGB video sequence—we create a consistent framework where the model treats every input and output as sequential frames, simplifying the multimodal learning process.

For each task, we insert a separator token ($\vert$) between the input and output frames to clearly delineate where the input ends and the prediction begins, guiding the model in distinguishing between these two parts of the sequence.

In tasks involving textual data, we use a simple tokenizer (splitting on spaces and also on punctuation) to break down the text into individual tokens, which we then render as video frames in the sequence. Each token is represented in a consistent format: a fixed-width font scaled to fill the $64 \times 64$ frame, ensuring clarity and compatibility with our video-based input structure. This approach allows the model to read and predict text as it would any other frame, effectively bypassing the need for traditional text-based tokenization while integrating text seamlessly with other modalities. By converting each token to a visual format, we enable cross-modal knowledge transfer, allowing the model to process text, images, and other modalities through a shared, frame-based learning paradigm.

\begin{figure}[h]
    \centering
    \includegraphics[width=0.9\linewidth]{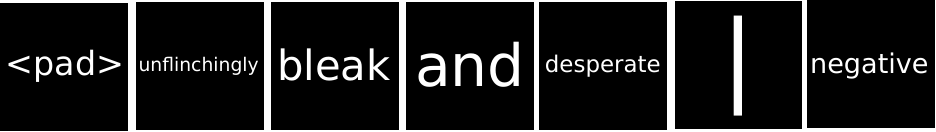}
    \caption{Truncated example of the SST2 sentiment dataset rendered as a video. Each square is a frame of the video sequence.}
    \label{fig:sst2-rendered}
\end{figure}

\paragraph{Text-to-Text} The SST2 (Stanford Sentiment Treebank 2) dataset is a widely-used benchmark for sentiment classification, consisting of thousands of movie review excerpts labelled with binary sentiment labels (positive or negative) \cite{socher2013recursive}. We use the text-encoding method described above, rendering each token as a video frame using a fixed-width font (Figure \ref{fig:sst2-rendered}).

\paragraph{Image-to-Text} As a simple test of image recognition ability, we employ the CIFAR-10 dataset. CIFAR-10 is a benchmark for image classification and consists of 60,000 color images, each with dimensions of $32 \times 32$ pixels \cite{krizhevsky2009learning}. These images are equally divided into ten distinct, mutually exclusive classes: airplane, automobile, bird, cat, deer, dog, frog, horse, ship, and truck.

\begin{figure}[h]
    \centering
    \includegraphics[width=0.5\linewidth]{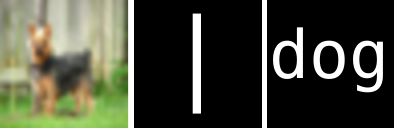}
    \caption{Truncated example of the CIFAR10 image classification dataset rendered as a video. Each square is a frame of the video sequence.}
    \label{fig:cifar10-rendered}
\end{figure}

In our reformulation approach, each CIFAR-10 image is resized to $64 \times 64$ pixels to fit our standardized input frame size. Following the resized image, we insert a separator frame, which visually signifies the end of the input and the beginning of the output sequence, followed by a frame containing the class label text rendered as an image (Figure \ref{fig:cifar10-rendered}).

\paragraph{Video-to-Text} We utilize the TinyVIRAT action classification dataset to rigorously assess the model’s capacity for understanding video content \cite{demir2021tinyvirat}. The TinyVIRAT dataset comprises of 7,663 training and 5,166 testing examples, each annotated with one or more of 26 distinct action labels that describe the activities depicted within low-resolution video sequences. This dataset is particularly noteworthy for its multi-label nature, allowing for nuanced action recognition; for instance, a video sequence might feature a man walking while carrying a backpack, which would be simultaneously labelled with both “walking” and “carrying” actions. This characteristic enables the evaluation of the model's ability to recognize and differentiate between overlapping actions occurring in dynamic environments.

\begin{figure}[h]
    \centering
    \includegraphics[width=1\linewidth]{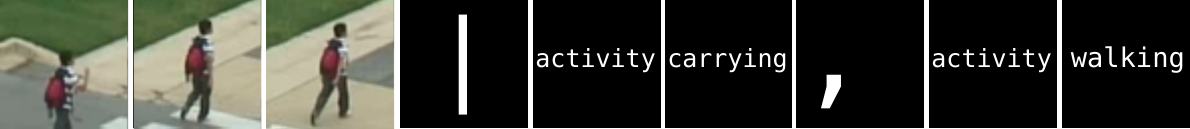}
    \caption{Truncated example of the Tinyvirat video classification dataset rendered as a video. Each square is a frame of the video sequence.}
    \label{fig:tinyvirat-rendered}
\end{figure}

To reformulate the task for next-frame prediction, we reformulate each video sequence by first presenting the input video frames, followed by a separator token, and concluding with a comma-separated list of action labels that correspond to the activities depicted in the video. (Figure \ref{fig:tinyvirat-rendered}).

\paragraph{Video+Text-to-Text} To explore the integration of multiple modalities within a single input, we leverage the CLEVRER (CoLlision Events for Video REpresentation and Reasoning) dataset, which is designed for video question answering (VQA) \cite{yi2019clevrer}. This dataset comprises synthetic video sequences depicting rendered 3D objects interacting through collisions, movements, and occlusions within a simple, controlled environment. Each video is paired with questions that require the model to understand and reason about these visual events. The questions in CLEVRER span various reasoning tasks, including descriptive, explanatory, and predictive questions about the events in the video, such as identifying specific objects, understanding object interactions, and predicting future states.

\begin{figure}[h]
    \centering
    \includegraphics[width=1\linewidth]{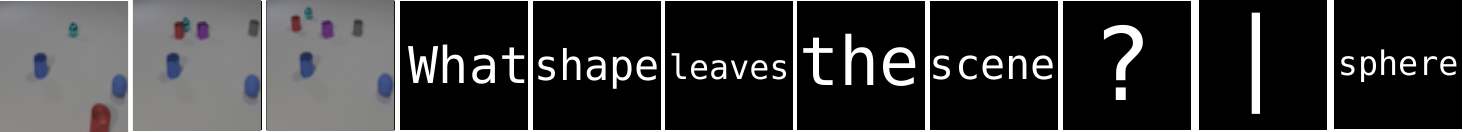}
        \caption{Truncated example of the CLEVRER task rendered as a video. Each square is a frame of the video sequence.}
    \label{fig:clevrer-rendered}
\end{figure}

In our framework, we represent each question and corresponding video as a unified sequence of frames. Each input sequence begins with frames of the CLEVRER video sequence subsampled to 4 frames. This is followed by the question text, separator token, and answer text encoded in the manner described previously (Figure \ref{fig:clevrer-rendered}). Similarly, we use the frame QA task from TGIF-QA dataset\cite{jang-IJCV-2019}, containing animated gifs with QA pairs.

To keep the sequence length manageable, we exclude complex counterfactual questions and multiple-choice answers from the dataset (only using the `descriptive' subset), as these would require additional frames and could lead to overly long sequences that complicate training.

\begin{figure}[h]
    \centering
    \includegraphics[width=0.9\linewidth]{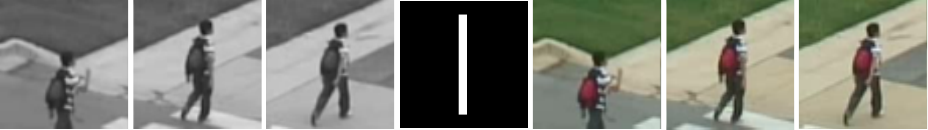}
    \caption{Truncated example of the colorization task rendered as a video. Each square is a frame of the video sequence.}
    \label{fig:colorization-rendered}
\end{figure}

\paragraph{Video-to-Video} We employ two tasks to test video-to-video performance. Firstly, we convert the TinyVIRAT videos into greyscale and test the model's ability to colourise the output. The sequence consists of the greyscale-converted video sequence, the separator frame, then the original RGB video sequence (Figure \ref{fig:colorization-rendered}).

\begin{figure}[h]
    \centering
    \includegraphics[width=0.9\linewidth]{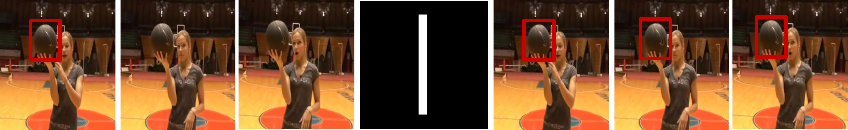}
    \caption{Truncated example of the LaSOT object tracking task rendered as a video. Each square is a frame of the video sequence.}
    \label{fig:lasot-rendered}
\end{figure}

Additionally, we test the model's ability to perform object tracking. The Large-scale Single Object Tracking dataset (LaSOT) consists of 1,550 video sequences hand-annotated with bounding boxes following an object of interest \cite{fan2019lasot}. We reformulate this task by converting it into a sequence which consists of the first frame of the video with the bounding box overlayed, the rest of the sequence, the separator frame, and finally the full sequence including the overlayed bounding box (Figure \ref{fig:lasot-rendered}).

\paragraph{Audio-to-Text} To demonstrate the ability to process audio data, we utilize the AudioMNIST dataset, which contains 30,000 audio recordings of spoken digits ranging from zero to nine \cite{becker2024audiomnist}. These recordings are spoken by multiple speakers with varying accents and intonations, providing a diverse dataset that challenges the model to recognize and classify spoken language across different vocal characteristics. 

\begin{figure}[h]
    \centering
    \includegraphics[width=0.5\linewidth]{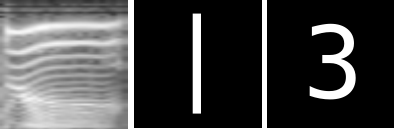}
    \caption{Example of the AudioMNIST audio classification dataset rendered as a video. Each square is a frame of the video sequence.}
    \label{fig:audio-rendered}
\end{figure}

In our framework, each audio sample is preprocessed into a spectrogram, which is used at the first frame in the video sequence, representing the audio content visually for consistency with other modalities. Following the spectrogram, a separator token frame is added, and the label digit is represented in the final frame as text (Figure \ref{fig:audio-rendered}). 

\begin{figure*}[t]
    \centering
    \includegraphics[width=0.8\linewidth]{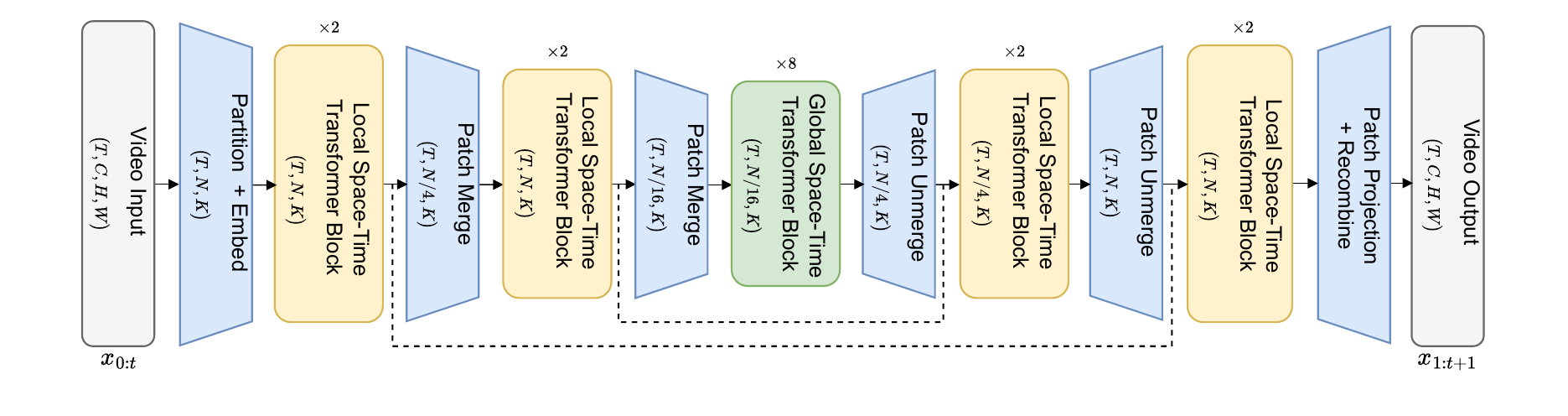}
    \caption{Illustration of the Transformer-based video prediction model used in this work. $T$ timestep video image frames are partitioned into $N$ non-overlapping patches and linearly embedded before being processed by a series of local and global space-time transformer blocks. Local blocks are restricted to local spatial and temporal attention windows, separated by non-overlapping patch merge/unmerge operations with a patch window size of 2. Global blocks operate on a smaller effective image resolution (e.g., for an input of 64 $\times$ 64 and a patch size of 8 $\times$ 8, the effective resolution at the global blocks is 16 $\times$ 16). Skip connections (dashed lines) are used between corresponding patch merge/unmerge operations to preserve information such as static background features.}
    \label{fig:model}
\end{figure*}

\subsection{Video Prediction Model}

We perform all tasks using a pure transformer-based model architecture. Building upon the successes of the ViT for image classification and TimesFormer \cite{bertasius2021space} for video understanding, we modify these architectures to accommodate for spatiotemporal modelling of video sequences and autoregressive video prediction. We further adapt the model to progressively process patches at smaller resolutions, similar to that of U-net style models, such that adjacent input patches are linearly merged to reduce the effective image resolution at each block on the encoding side. This design choice allows us to train end-to-end models efficiently without resorting to pretrained encoder/decoder models. As illustrated in \cref{fig:model}, partitioning into patches is followed by a series of local space-time transformed blocks separated by patch merge operations, with a global space-time transformer block operating on the reduced resolution patch-based representations. This has the benefit of capturing multiscale spatial features, while reducing the computational complexity of the global self-attention operations due to a smaller number of image patches. The reverse process is applied on the decoding side of the model to produce a predicted next image frame for each input frame. During inference, the final predicted frame is concatenated to the input sequence in an autoregressive manner. We outline the key model components and considerations below. Our model uses 41M parameters (for comparison, Transframer \cite{nash2022transframer} uses 470–829M parameters). 

\paragraph{Patch Representation}
A core feature of the model is in the partitioning of input video frames into non-overlapping patches, similar to ViT. These patches are processed using local and global spatiotemporal self-attention mechanisms as described in \cite{bertasius2021space}, with spatial self-attention focusing on intra-frame patch relationships and temporal self-attention handling inter-frame causal dependencies. Temporal causality is enforced by masking future patches during training, ensuring that only past and present information is available for prediction. The model takes as input a video sequence $\mathcal{X} = \{x_t\}_{t=1}^T$, such that each $x_t \in \mathbb{R}^{C \times H \times W}$ is a $C$-channel image frame of height $H$ and width $W$ at timestep $t$ for $t = 1, 2, \ldots, T$. Frames are partitioned into $N$ equally-sized non-overlapping patches, and linearly embedded to a 1D representation of size $K$ (we set $K$ to 512 throughout). This embedding is combined with learnable spatial and temporal positional encodings to form the input sequence for the transformer. Spatial attention is applied between image patches within each frame independently, while causal temporal attention is applied across timesteps at the same spatial patch location.

\subsection{Training Details}

\begin{table}[]
    \centering
    \scalebox{0.9}{
    \begin{tabular}{lccc}\toprule
    Task                 & Input Len & Target Len & Batch Size \\\midrule
    Text Classification  & 21        & 1          & 16           \\
    Image Classification & 2         & 1          & 128           \\
    Video Classification & 11        & 1          & 16           \\
    Audio Classification & 2         & 1          & 256        \\
    Video QA             & 21        & 1          & 16           \\
    Object Tracking      & 21        & 20         & 8          \\
    Video Colorization   & 11        & 10           & 16          \\\bottomrule
    \end{tabular}}
    \caption{Training configurations for each task}
    \label{tab:params}
\end{table}

We train our model on each dataset independently, with no language model or image pretraining, in order to determine the feasibility of our task reformulation approach with no latent performance benefits from transferring existing learned representations. We train end-to-end with a Multi-Scale Structural Similarity Index Measure (SSIM) loss \cite{1284395}, with a constant learning rate of $3 \times 10^{-4}$, and a batch size of 8-32 depending on the task context length. We train with the AdamW \cite{Loshchilov2017DecoupledWD} optimizer using default parameters, and set dropout to 0.1 for all layers except the final output layer. All models are evaluated on checkpoints corresponding to the best SSIM validation performance, with all models trained on a single NVIDIA A100 for a maximum of 7 GPU days. The input/output lengths and batch sizes vary per task, and are detailed in Table \ref{tab:params}. 

\section{Experiments}

\begin{table*}[h]
    \centering
    \resizebox{0.95\textwidth}{!}{%
\tabcolsep=0.11cm
\begin{tabular}{lllccccccc}\toprule
Task                 & Dataset     & Model                                   & PSNR ↑  & SSIM ↑ & F1 ↑ & Acc ↑ & IoU ↑ & CDC ↓ & Colorfulness ↑\\ \midrule
Text Classification  & SST-2       & Ours                                    & 41.8     & 98.7  & 76.8   & 75.5     &   -   &  -   &  -   \\
                     &              & BERT-base                              & -        & -     & -      & 81.2     &   -   & -    &  -    \\
                     &              & Vega v1 \cite{zhong2023bag}\textdagger & -        & -     & 91.3   & -         &   -   & -    &  -    \\\midrule
Image Classification & CIFAR-10    & Ours                                    & 45.7     & 95.9  & 89.1   & 89.1     &   -   & -    &  -    \\
                     &              & ViT-base                               & -        & -     & 71.3   & 71.3     &   -   & -    &  -    \\
                     &              & Resnet-101                             & -        & -     & 90.0   & 90.0     &   -   & -    &  -    \\
                    &              & PCANet \cite{chan2015pcanet}            & -        & -     & 77.1   & 77.1     &   -   & -    &  -    \\
                     &       & ViT-huge \cite{dosovitskiy2021an}\textdagger  & -        & -     & 99.5   & 99.5     &   -   & -    &  -    \\\midrule
Video Classification & TinyVIRAT   & Ours                                    & 50.1     & 97.3  & 74.1 (30.4*)  & 60.4     &   -   & -    &  -     \\
                     &              & ResNet50\textdagger                    & -        & -     & 29.1*   &  -       &   -   & -    &  -    \\
                     &              & WideResNet\textdagger                  & -        & -     & 32.6*   &  -       &   -   & -    &  -    \\\midrule
Audio Classification & audioMNIST  & Ours                                     & 50.9     & 98.9  & 96.9   & 97.1     &   -   &  -   &  -     \\
                     &             & AlexNet \cite{becker2024audiomnist}      & -        & -     & -      & 95.8       &   -   & -    &  -    \\\midrule
Video QA             & CLEVRER     & Ours                                     & 25.7     & 79.8  & 52.4   & 52.5     &  -   &   & -      \\
                     &             & LSTM \cite{yi2019clevrer}\textdagger     & -        & -     & -      & 34.7       &   -   & -    &  -    \\
                     &             & + CNN \cite{yi2019clevrer}\textdagger    & -        & -     & -      & 51.8       &   -   & -    &  -    \\
                     & TGIF-QA        & Ours                                  & 42.0     & 98.5  & 53.2   & 52.3          &  -   &  -  & - \\
                     &              & LSTM + Attn \cite{jang-IJCV-2019}       & -        & -     & -      & 51.9          &  -   &  -  & - \\\midrule
Object Tracking      & LaSOT       & Ours                                     & 35.7     & 98.7  & -      & -     & 0.63  &  -   & -      \\
                     &             & \citet{Dai_2020_CVPR}                    & -        & -     & -       & -    & 0.60  &  -  & -       \\\midrule

Video Colorization   & TinyVIRAT   & Ours                                     & 42.4    & 99.7    & -       &  -    & -     & 0.02   & 73.1   \\
                     &             & \citet{zhang2016colorful}                & -       & -       & -       & -     & -     & 0.02  & 82.2 \\
                     &             & \citet{yang2024colormnet}                & -       & -       & -       & -     & -     & 0.03   & 72.0 \\\bottomrule
\end{tabular}}
    \caption{Performance metrics across various tasks using our proposed next-frame prediction framework. Metrics include Peak Signal-to-Noise Ratio (PSNR), Structural Similarity Index (SSIM), F1-score and Accuracy (F1, Acc), Intersection over Union (IoU), Color Distribution Consistency index (CDC), and Colorfulness, which are each tailored to evaluate specific task outputs. *macro F1-score to allow direct comparison with baseline models. \textdagger indicates models with pretraining on other datasets.}
    \label{tab:main_results}
\end{table*}

We train on each task in turn in a single-task manner (we leave training on all the tasks jointly in a multitask setting as future work). For tasks which output text, we perform OCR on the output frames using the open-source tesseract OCR engine\footnote{\url{https://github.com/tesseract-ocr/tesseract}}. For tasks which have a fixed vocabulary (e.g. classification tasks), the OCR text is matched to the closest vocabulary word to minimise any OCR-related errors. For classification tasks, the resulting text is measured using F1-score and Accuracy.

The object tracking task is evaluated by extracting the bounding box in the outputted frames, then comparing these to the labelled boxes using Intersection over Union (IoU).

For video colorization, we evaluate both colorization quality and temporal consistency. This dual evaluation is crucial, as a model could appear consistent by merely reproducing the grayscale input. We use PSNR and the colorfulness metric from \citeauthor{liu2024temporally} to assess color diversity. Temporal consistency is measured using the Color Distribution Consistency (CDC) index \cite{liu2024temporally}, which calculates the Jensen-Shannon (JS) divergence of color distributions across frames:

$$
CDC_t = \frac{1}{3 \times (N - t)}\sum_{c \in \{r, g, b\}}{\sum_{i=1}^{N-t}{JS(P_c(I^i), P_c(I^{I+t}))}}
$$

where $P_c(I^i)$ is the normalized probability distribution of channel $c$ in frame $I^i$ of $N$ total frames. Overall consistency is computed as:
$$
CDC = \frac{1}{3} (CDC_1 + CDC_2 + CDC_4)
$$
\section{Results}

The results across each task are presented in Table \ref{tab:main_results}, while Figure \ref{fig:sample-outputs} illustrates several sample outputs produced by our model, offering a qualitative perspective on its performance.

For the SST-2 text classification task, our model achieves an F1-score of 76.8. Although this is lower than the current state-of-the-art performance of 91.3, that was achieved through the use of a ``Bag of Tricks'' methodology involving extensive pretraining and complex finetuning processes \cite{zhong2023bag}. Our model is simply trained on the SST-2 dataset, with no use of pretrained weights or advanced external embeddings, making it comparable to other models used without pretraining (See the BERT-base result). Some portion of this lower score can also be attributed to truncating any inputs over 20 tokens (limiting the evaluation to shorter inputs 20 tokens or fewer increases the F1-score to 80.0).

On the CIFAR-10 image classification task, our model attains an accuracy of 89.1. While this score is lower than fine-tuned ViT models leveraging pretrained weights, which reach an impressive accuracy of 99.5 \cite{dosovitskiy2021an}, our model performs significantly better than simpler baseline models, such as PCANet, which achieves 77.1 accuracy \cite{chan2015pcanet}. This performance suggests that our reformulation approach is able to perform image classification tasks without any pretraining, surpassing older, simpler architectures while trailing behind state-of-the-art models that benefit from extensive pretraining.

In multi-label image classification on the TinyVIRAT dataset, our approach successfully outputs multiple labels per instance. When compared to baselines, our model surpasses the ResNet50 baseline (29.1) in performance but does not reach the level of the WideResNet model (32.6) \cite{demir2021tinyvirat}.

On the AudioMNIST audio classification task, the model achieves an accuracy of 97.1, surpassing the AlexNet baseline score of 95.82 from the original dataset paper \cite{becker2024audiomnist}. This notable improvement demonstrates the model’s efficacy in audio-based classification tasks. Analysis of misclassifications reveals that the most frequent error involves confusion between the spoken digits “four” and “five,” suggesting areas for potential refinement in distinguishing phonetically similar sounds.

For the CLEVRER VQA (Visual Question Answering) dataset, our model achieves an accuracy of 52.5 on descriptive questions. This performance surpasses both the LSTM (34.7) and LSTM+CNN (51.8) baselines, illustrating our model’s ability to understand the relationship between visual and textual information without separate encoder-decoder models \cite{yi2019clevrer}. It is important to note that due to hardware limitations, and the long question lengths included in the CLEVRER dataset, we subsampled the video to only include 4 frames of the moving scene, while also reducing the resolution to 64 $\times$ 64, significantly limiting the accuracy achievable by our approach.

In the LASOT object tracking task, the model reaches an intersection over union (IoU) of 0.63, demonstrating a consistent ability to track objects throughout video sequences. Analysis of tracking outputs reveals that the model effectively follows the object of interest across frames with accurately drawn bounding boxes. However, tracking accuracy diminishes slightly toward the end of longer sequences, with the borders of the overlaid bounding box eroding, likely due to autoregressive pixel-level error propagation.

On the colorization task, our model shows improved color diversity in the generated outputs compared to the ground truth (73.1 color diversity for our model versus 70.6 in the original dataset). However, it exhibits less consistency in color application across the video sequence (0.0169 consistency for our model versus 0.00522 in the original dataset). This trade-off between diversity and consistency suggests that while our approach introduces a more vibrant color palette, further refinement could improve temporal coherence, particularly for applications requiring uniform color continuity across frames.

\begin{figure*}[t]
    \centering
    \vspace{-1.5em}
    \includegraphics[width=0.72\linewidth]{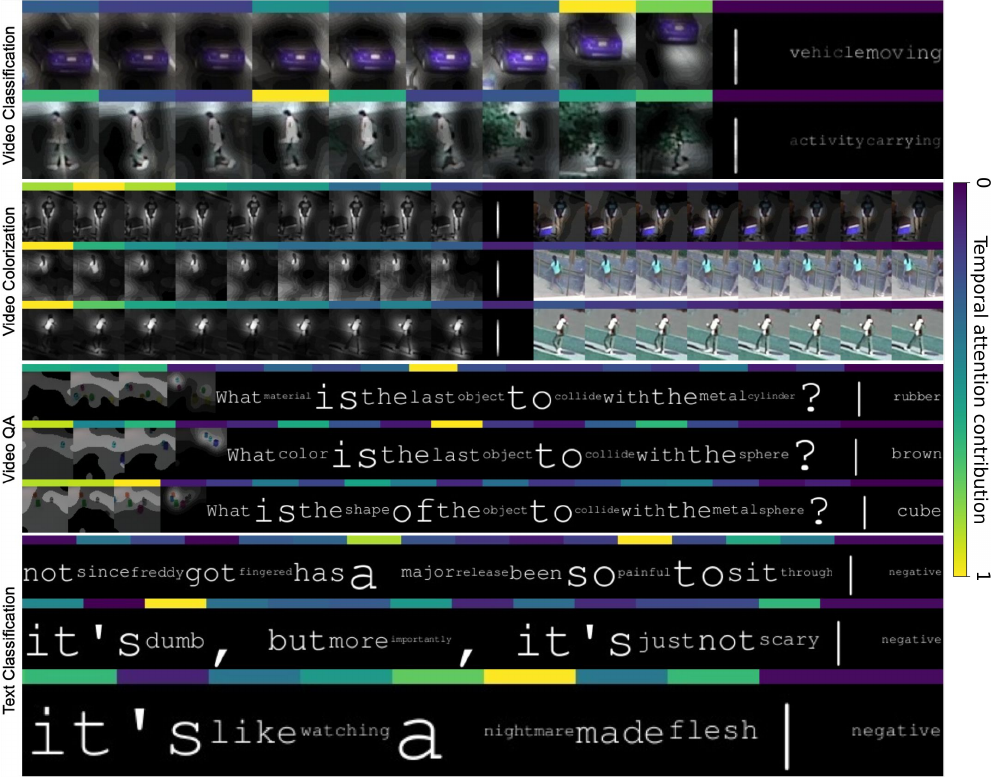}
    \vspace{-0.5em}
    \caption{Attention visualisations showing where the model attends to both spatially (indicated by light/dark areas overlaid on the frame), and temporally (indicated by a colour scale above the sequence). Spatial attention scores are calculated by taking the patch-wise contribution per frame, averaged over all layers. Temporal attention scores are calculated by taking the temporal attention weights for each input frame when generating the target frames, averaged over all global space-time attention layers.}
    \label{fig:attn_1}\vspace{-1.5em}
\end{figure*}

ch\subsection{Attention Maps}

In order to examine what is being learned by our model, we visualize internal patch-wise attention scores both spatially (within a frame) and temporally (between frames) in Figure \ref{fig:attn_1}.

In the TinyVIRAT video classification task, the model pays most attention to the object/person performing the action, as well as frames and pairs of frames which indicate the class. In the first example (top row), spatial attention focuses on image patch locations containing the edges of the car which suggest relative movement, and temporal attention is highest for the final two frames in which the vehicle moves position, indicating the label ``vehicle moving'' should be output. 

For colorization, the greatest attention is paid to the first frames of the input sequence, which we suggest allows the model to produce consistent coloring across the whole sequence (more attention is also paid to the first frame of the output sequence, presumably to maintain color consistency). Spatially, the model clearly attends to key moving objects and areas of distinct color. 

In the CLEVRER examples, we see how the model is able to focus on question frames which contains key words needed to produce the output, such as ``color'' and ``metal''. Within the moving object scene frames, we see how the model spatially attends to objects and their trajectory paths, correctly identifying objects such as ``the metal cylinder'' and ``the sphere''.

Lastly, for sentiment text classification, we observe a similar pattern to attention maps in LLMs, where the model pays more attention to words with a strong emotive sentiment. In our model we see frames containing the words `nightmare', `painful', and `dumb' exhibiting high temporal attention weightings, as they are strong indicators of the overall sentiment of the review.

\subsection{Limitations and Future work}

The aim of this work is not necessarily to surpass state-of-the-art performance on any of the tasks, but to show that unifying these modalities via reformulation is possible, without heavily compromising on performance relative to task-specific baselines. Future work should aim to scale this up, improving performance and tackling more complex tasks. We suggest that this can be done by mirroring the improvements gained from pretraining in NLP and computer vision by first training the model on unstructured tasks from each modality. Tasks such as language modelling and image captioning can be naturally reformulated as next-frame-prediction and included in our paradigm. This is especially important as a large portion of the current training time on tasks such as classification and question answering is dedicated to learning to output words rather than solving the task itself. Beginning with a checkpoint already trained on language modelling would mitigate this.

\vspace{-0.4em}
\section{Conclusion}

In the age of large foundation models, the trend is towards larger, more unified, fully end-to-end models without task-specific components. We introduced a novel multimodal learning framework that reformulates various multimodal tasks into a unified next-frame prediction paradigm. This approach addresses critical limitations in current multimodal model designs, which often require modality-specific encoders and are limited in scalability and flexibility when adapting to new tasks. By unifying diverse multimodal tasks under a single framework, our model can handle text, image, audio, and video inputs without modality-specific components, significantly simplifying the design and training process. We show it is possible to train a simple next-frame prediction model to solve these tasks, with many of them matching the performance of comparable models trained without pretraining data.

Ultimately, our work lays the groundwork for building more generalized and efficient multimodal foundation models, towards systems that can process information from diverse modalities in a unified and scalable way.

{
    \small
    \bibliographystyle{ieeenat_fullname}
    \bibliography{main}
}

\end{document}